\PassOptionsToPackage{table}{xcolor}
\documentclass[runningheads]{llncs}

\usepackage{eccv}

\usepackage{eccvabbrv}

\usepackage{graphicx}
\usepackage{booktabs}
\usepackage[mathscr]{eucal}
\usepackage[accsupp]{axessibility}  % Improves PDF readability for those with disabilities.
\usepackage{paralist}
\usepackage{multirow}
\usepackage[table]{xcolor}

\usepackage[pagebackref,breaklinks, citecolor=gray]{hyperref}
\hypersetup{hidelinks} 

\usepackage{orcidlink}

\begin{document}

% ---------------------------------------------------------------
% TODO REVIEW: Replace with your title
\title{Spatial-Temporal Multi-level Association for Video Object Segmentation} 

% TODO REVIEW: If the paper title is too long for the running head, you can set
% an abbreviated paper title here. If not, comment out.
\titlerunning{Spatial-Temporal Multi-level Association for Video Object Segmentation}

% TODO FINAL: Replace with your author list. 
% Include the authors' OCRID for the camera-ready version, if at all possible.
\author{Deshui Miao\inst{1*} \and
Xin Li\inst{2*}\ \and
Zhenyu He\inst{1 \lozenge}\ \and
Huchuan Lu \inst{3}\ \\ \and
Ming-Hsuan Yang \inst{4,5,6} }

% TODO FINAL: Replace with an abbreviated list of authors.
\authorrunning{Deshui Miao et al.}
% First names are abbreviated in the running head.
% If there are more than two authors, 'et al.' is used.

% TODO FINAL: Replace with your institution list.
\institute{Harbin Institute of Technology, Shenzhen  \and
Peng Cheng Laboratory\\ \and
Dalian University of Technology\\ \and
University of California at Merced\\ \and
Yonsei University\\ \and
Google Research }

\maketitle

\begin{abstract}
   Existing semi-supervised video object segmentation methods either focus on temporal feature matching or spatial-temporal feature modeling.
However, they do not address the issues of sufficient target interaction and efficient parallel processing simultaneously, thereby constraining the learning of dynamic, target-aware features.
To tackle these limitations, this paper proposes a spatial-temporal multi-level association framework, which jointly associates reference frame, test frame, and object features to achieve sufficient interaction and parallel target ID association with a spatial-temporal memory bank for efficient video object segmentation.
Specifically, we construct a spatial-temporal multi-level feature association module to learn better target-aware features, which formulates feature extraction and interaction as the efficient operations of object self-attention, reference object enhancement, and test reference correlation.
In addition, we propose a spatial-temporal memory to assist feature association and temporal ID assignment and correlation.
We evaluate the proposed method by conducting extensive
experiments on numerous video object segmentation datasets, including DAVIS 2016/2017 val, DAVIS 2017 test-dev, and YouTube-VOS 2018/2019 val.
The favorable performance against the state-of-the-art methods demonstrates the effectiveness of our approach.
All source code and trained models will be made publicly available.
\end{abstract}

\section{Introduction}
\label{sec:intro}
% xin:
% The logic of introduction
% P1: Introduction of the VOS task, its application, and its challenges.
% P2: How do existing VOS methods handle these challenges and why they cannot address them well
% P3: The core insights: the key points to solve the above challenges.
% P4: How we solve these challenges. Such as, To this end, we xxx

Semi-supervised Video Object Segmentation (VOS) aims to delineate and track the objects specified by the given masks within a video sequence~\cite{hvos, vos_adap, ytvos2018, ISVOS, STM, onlinevos}. 
This practice holds significant promise across various applications, particularly as the prevalence of video content surges in domains like autonomous driving, augmented reality~\cite{soe, joint,psg}, and interactive video editing~\cite{vos-vfi, psg, vpsg}.
%
% xin: I added a sentence to show the challenges of VOS.
%
The core challenge of VOS lies in utilizing the limited initial information (one target sample) to accurately track and segment the target object that undergoes various visual changes and interacts with a complex video environment.

\begin{figure*}
  \centering
    \includegraphics[width=0.98\linewidth]{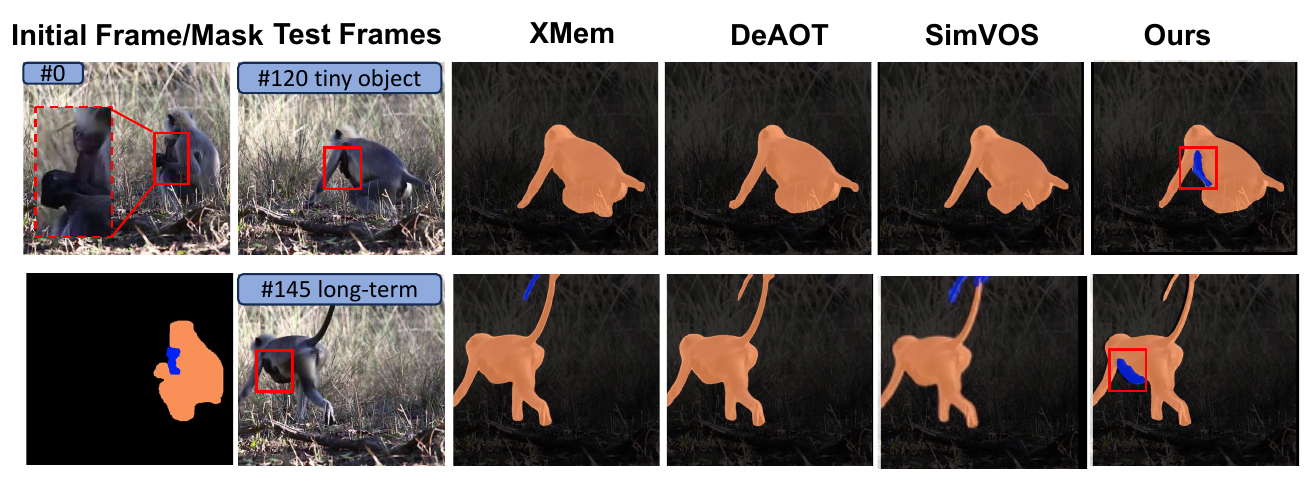}
   \caption{\textbf{Performance on challenging VOS scenarios with tiny objects and long-term changes.} XMem~\cite{xmem}, DeAOT~\cite{deaot}, and SimVOS~\cite{SimVOS} do not work well on this scenario. Our method accurately predicts the mask of the `baby monkey' (marked by the red box) over frames.
   }
   \label{fig:first-fig}
\end{figure*}

% xin: one kind of VOS methods focus on learning more comprehensive asssicaition features between the example and target sample.
% xin: Another explore model temporal information to handle target variations.
% xin: at last, give the weakness of existing methods, such as do not take these two jointly, or does work well on mining temporal or spatial information.
%%%%%%%%%%%%%%%%
% mds: 
Early VOS methods primarily involve fine-tuning segmentation models~\cite{caelles2017oneshot, onlinevos, fastonline} on annotated videos or developing pixel-wise matching maps~\cite{fastvos, feelvos}.
Recent approaches~\cite{STM, xmem, aot, deaot, stcn} mainly rely on matching-based frameworks, which %focus more on temporal information and 
build associations between the test frame features and memorized target features to %handle target variations.
infer the target state.
The matching-based methods enable efficient temporal ID association by calculating the similarity between memorized features and the test feature, facilitating target association across multiple frames, and efficiently handling multi-object scenarios.
%
% However, since the temporal information association process is performed after feature extraction, losing target details during this phase yields challenging segmentation results for small or faint objects.
% xin: you may merge this paragraph into the former one or not based on the length of these two paragraphs.
To better model dynamic target feature learning, number VOS methods~\cite {SimVOS, vos_corr} explore integrating feature association into the feature extraction process to achieve a more comprehensive spatial feature correlation between reference and test frames.
However, temporal feature association, which associates features after feature modeling, results in ineffective feature association when dealing with small object segmentation. 
Meanwhile, the method of spatial feature association suffers from modeling long-term associations.
% However, since simultaneous feature extraction and feature association require substantial computational resources, and due to the lack of an efficient long-term temporal ID propagation module, this approach processes one target at a time before aggregating the results.
%
% Although SimVOS~\cite{SimVOS} introduces a token refinement module to enhance model efficiency, it does not address the issue of multi-target ID propagation.

% xin: The core points in comprehensive target feature learning: 1) spatial information to ensure accurate prediction for small objects, 2) temporal consistency that maintains the object's identity to distinguish a target from other objects and the backgrounds. 
% The challenges in developing a VOS method that simultaneously achieves comprehensive feature association and efficient temporal multi-target ID assignment are twofold: 1) How can information processing be streamlined to enable more comprehensive and efficient spatial-temporal interaction? 2) How can various IDs be effectively propagated to enrich target information in feature extraction and ensure more accurate ID assignments?
% 2) How can different IDs be propagated effectively to enrich the target information in the feature extraction and assign the ID more accurately?
The core points in comprehensive target feature learning are: 1) spatial information to ensure accurate prediction for small objects, and 2) temporal consistency that maintains the object's identity to distinguish a target from other objects and backgrounds. 
The above two challenges motivate us to design an efficient and sufficient spatial-temporal multi-level association method for the VOS task.
%

% xin: update this paragraph according to the new framework
In this paper, we propose a spatial-temporal multi-level association (STMA) framework comprising a spatial-temporal multi-level feature association module (STML), a prediction part, and a spatial-temporal memory bank.
During the feature extraction stage, object features, reference frames from the spatial memory, and the test frame are input into the STML module for asymmetric information interaction.
%
% The key to the STML is to enrich the target information in multi-level layers, where the object information flows to references for target enhancement and then to the target feature through references for temporal feature association.
%
Specifically, in the STML module, we decouple the self-attention into three substreams: object feature self-attention, reference object enhancement, and test reference correlation.
The STML module can then learn dynamic target-aware features.
Equipped with these features, we then perform object-related semantic matching and ID correlation with the temporal memory bank.
%
% The ID propagation module primarily consists of the memory bank and the ID propagation process.
%
The temporal memory bank stores historical features related to the different targets, which serve keys and values for ID assignment. 
The ID correlation module matches the target features with their respective IDs. 
It completes the transformation from one feature map to multiple target-related feature maps, ensuring each feature map contains a target with a unique ID.
Especially, in the STML module, the target feature is comprehensively learned without differentiating between the IDs of different targets. In ID association, the IDs of different targets are obtained by comparing them with the IDs in temporal memory.
We conduct extensive experiments on various public VOS datasets, including DAVIS 2016 \& 2017 and YouTube-VOS 2018 \& 2019.
The favorable performance against the state-of-the-art methods on all these datasets demonstrates the effectiveness of the proposed algorithm, especially in handling challenging sequences with small targets or long-term duration.

The main contributions of this work are:
\vspace{2mm}
\begin{compactitem}
 \item We propose a spatial-temporal multi-level feature association module to facilitate efficient spatiotemporal target information interchange for video object segmentation.
\vspace{2mm}
 \item We develop a spatial-temporal memory bank to assist the STML module and ID assignment for long-term modeling.
 It retains information pertinent to the targets from previous frames, which is utilized to match, segregate, and enhance the features of each target in the test frame.
\vspace{2mm}
 \item We conduct extensive experiments to demonstrate the effectiveness of the proposed method.
 Our method achieves favorable performance against other state-of-the-art methods on extensive VOS datasets.
 Specifically, Our method gains significant improvement on DAVIS 2017 val (88.9\% $\mathcal{J}\&\mathcal{F}$), DAVIS 2017 test (85.6\% $\mathcal{J}\&\mathcal{F}$) and YouTube-VOS 2019 (86.3\% $\mathcal{J}\&\mathcal{F}$) without any pre-training.
\end{compactitem}

\section{Related Work}
\label{sec:relatedWork}
% 从matching based和memory 建模两个方面去说这个related works。
% We discuss the closely related VOS methods which are grouped into two categories based on whether they adopt a separate learning approach or a joint learning method.
% xin: this section should be reorganized according to the following ways
% 1) group existing works as a) spatial feature learning; b) temporal information modeling (similar the Memory construction on VOS part); c) joint feature learning (if there is not any VOS methods belonging to this category, then you can review a few works on other video object perception task, such as tracking. And you can point out the challenge that it is time-consuming for VOS method to do this directly)
% 2) The current version is with less logic. If you want to review several kinds of methods, you may first write 'there are several kinds of xxx methods, including a,b,c'. 
% Note that if these kinds of methods have a relationship (such as, they solve a problem of another kind, then you can introduce them in a chain.)
%%%add a total description
We discuss the closely related methods from the aspects of feature association and memory construction.

\vspace{1mm}
\noindent\textbf{Feature Association in VOS.}
Numerous methods~\cite{caelles2017oneshot,onlinevos, fastonline,maskrnn, premvos, motioncrn, transductive, globalins, swiftnet, calibration} employ online learning at test time to update models and build connections between consecutive frames to transmit segmentation masks sequentially, which is not efficient in the fine-tuning step.
Although some methods~\cite{osmn, pml} improve the test efficiency, they are often susceptible to cumulative errors due to occlusions or tracking drift. 
Similar to STM~\cite{STM}, recent methods focus more on memory matching and propagation with a feature memory bank and matching strategy.
In this pattern, feature matching and propagation are performed after feature extraction.
XMem~\cite{xmem} meticulously crafts distinct memory storage mechanisms and proposes the Atkinson-Shiffrin model for matching, yielding impressive results.
Zhang et al.~\cite{vos_corr} develop a correspondence-aware training framework that enhances propagation-based VOS solutions by explicitly fostering robust correspondence matching throughout the network learning process.

In addition, other approaches~\cite{aot, deaot, lattnvos} leverage transformers for feature interaction and spatiotemporal information propagation.
Aiming to achieve simultaneous multi-object tracking and enhance transformer efficacy within VOS, AOT~\cite{aot, deaot} proposes an ID propagation mechanism and advances the Long-Short-Term-Transformer (LSTT)~\cite{lstt} for VOS and achieve promising performance.
%
%
% DeAOT~\cite{deaot} boosts the performance of AOT by decoupling feature matching from ID propagation.
%
However, the aforementioned methods, performing temporal feature matching after feature extraction, do not work well in handling small and faint objects, since the fine-grained details can be dismissed during the feature extraction process.
Similar to extensive single object tracking methods~\cite{ostrack, SimTrack,mixformer}, SimVOS~\cite{SimVOS} performs the joint extraction and association of target features between historical and current frames in ViT blocks.
However, SimVOS processes every target sequentially during the training and testing phases, which is time-consuming and does not consider the long-term temporal association between the targets.
% 兼顾corse to fine,两个memory，两种缺陷不同。
\begin{figure*}
  \centering
   %\fbox{\rule{0pt}{2in} \rule{0.9\linewidth}{0pt}}
   \includegraphics[width=1.0\linewidth]{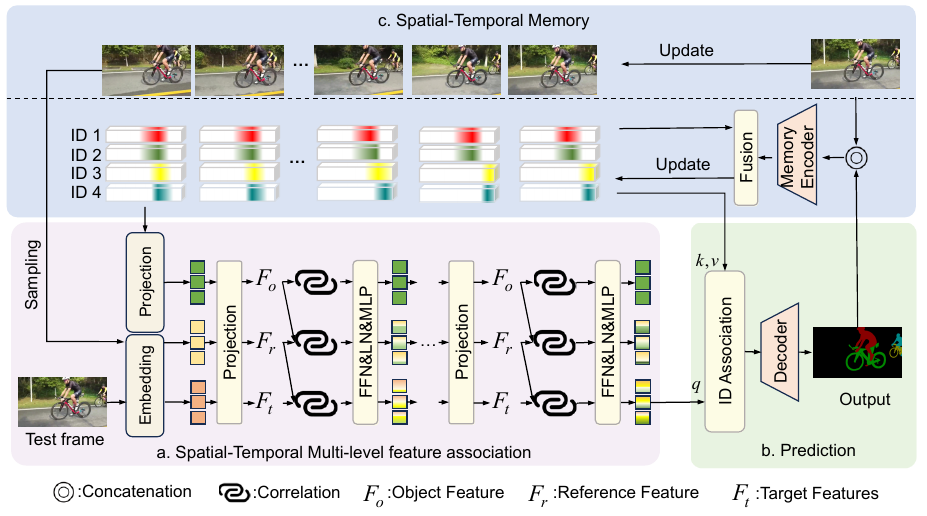}
   \caption{\textbf{Overall framework.} It consists of a spatial-temporal multi-level (STML) feature association part, a prediction module, and a spatial-temporal memory. The STML module conducts simultaneous feature extraction and correlation. The spatial-temporal memory not only provides object features and reference frames for STML but also offers temporal feature information for ID association..}
   \label{fig:overall}
\end{figure*}

\vspace{1mm}
\noindent\textbf{Memory construction in VOS.}
Recent advancements in the field have focused on incorporating a larger number of past frames into a feature memory bank to overcome limitations related to context understanding. 
STM~\cite{STM} gains significant attention for this purpose and has inspired a range of subsequent research. 
However, a common challenge with most STM variants~\cite{stcn,xmem,vos_corr} is their struggle to process long videos without causing an excessive increase in the size of the feature memory bank.
Although AOT~\cite{aot, deaot} introduces an adaptation of the attention mechanism for transformers, it does not address the issue of GPU memory overload. 
Meanwhile, some approaches~\cite{joint,globalins} implement a localized feature memory window, which unfortunately overlooks the importance of capturing long-term context beyond this limited scope.
To address the long-term VOS problem, XMem~\cite{xmem} designs an updating architecture that integrates several independent but intricately linked feature memory stores.
SimVOS~\cite{SimVOS} facilitates information transfer between different frames by updating reference frames.
To better model spatial-temporal information, we construct spatial-temporal memory and perform spatial-temporal multi-level feature association within a single framework to achieve both efficient feature representation and effective ID association.
Furthermore, we introduce object features to enhance the discriminability between different target representations, thereby better distinguishing between multi-targets and enhancing the accuracy of ID association.
%
% We synthesize the efficiency of separated feature learning and the sufficient association of joint feature learning approaches, proposing a tri-flow network for efficient and comprehensive information extraction and propagation.
% %
% Unlike SimVOS, which concatenates masks directly onto reference frames, our proposed method uses object features to represent targets. 
% %
% This approach allows us to augment targets within reference frames, offering the flexibility to handle multiple targets effectively.
% %
% It is worth mentioning that in the block of tri-flow association, we merely enhance the features of the targets without distinguishing them. 
% %
% Upon completing feature extraction and feature association in the tri-flow association, we model and segregate each target feature with short-term and long-term memory for effective ID propagation.

\section{Proposed Algorithm}
% 达到什么目的，做了什么。目标：介绍目标，为了设计这个目标，设计了什么模块。
% 
%Our approach aims to learn a comprehensive spatial-temporal target representation and formulate the ID association of multiple targets across frames efficiently for video object segmentation.
% xin: the above sentence does not clearly describe what you do. Think!
%
Our approach aims to learn a spatially-aware and dynamically-distinctive target representation for video object segmentation.
To this end, we propose a spatial-temporal multi-level association VOS framework comprising a spatial-temporal memory to remember and update the target object information adaptively, a feature association part to generate target features, and a prediction component to generate the final results.

\subsection{Overall Framework}
\label{Overallframework}
% efficient triplit association
%As shown in Figure (), our baseline model mainly consists of a joint feature extraction and ID embedding block, memory management, and a mask prediction decoder. 
%We use the ConvViT as our backbone mainly because: 1) ConvViT can not only perform joint feature extraction and interaction but also multi-scale feature generation, which perfectly meets our design for VOS; 2) ConvViT has been pre-trained by MAE method, which can boost the performance of various downstream tasks.

Given the test frame $\mathbf{X}_{t}\in \mathbb{R}^{3 \times H \times W}$ with $n$ target objects, the goal of our method is to predict the target masks $\mathbf{M}_{t}\in \mathbb{R}^{H \times W}$ of $\mathbf{X}_{t}$ based on the given reference frames $\mathbf{X}_{r}\in \mathbb{R}^{m \times 3 \times H \times W}$, where $m$ is the number of reference frames.
%
%Rather than feeding all the test and reference frames directly into the ViT~\cite{vit} backbone for combined feature extraction and interaction, they are initially transformed into input sequences.
The proposed method takes the test frame feature, reference frame feature, and object feature as input.
The features of the test and reference frames are generated using a linear model.
Specifically, we convert each input frame into a string of 2D flat patches with dimension $N \times 3P^2$, resulting in the reconfigured reference sequences, namely, $\mathbf{X}_{r}\in \mathbb{R}^{m \times N \times 3P^{2}}$ and the test sequence $\mathbf{X}_{t}\in \mathbb{R}^{N \times 3P^2}$, where $P^2$ is the patch size and $N=HW/P^2$ is the number of patches.
After employing the linear projection $\mathbf{E}\in \mathbb{R}^{3P^2 \times C}$ to transform the 2D patches into 1D tokens of $C$ dimensions and incorporating the sinusoidal positional embedding $\mathbf{P}\in \mathbb{R}^{N \times C}$, we get the reference feature $\mathbf{F}_r\in \mathbb{R}^{m \times N \times C}$, target feature $\mathbf{F}_x\in \mathbb{R}^{N \times C}$.
The object features $\mathbf{F}_{o}\in \mathbb{R}^{n \times C}$ are generated using a projection layer and a position embedding operation.
%
% For online updates of the object ID embedding, we propose a short-term memory to conduct the update during training and inference.
%

Upon the features of the three inputs, the proposed method conducts spatial-temporal multi-level feature association to generate the correlated target features by leveraging the attention-based correlation.
% To model the correspondence of three different components, we propose to perform propagation with detailed target information.
We then split the target feature from the above-generated feature and perform ID association to get the feature map of every target.
Specifically, we calculate the query-key affinity matrix between $\mathbf{F}_t$ and $\mathbf{Fk}\in \mathbb{R}^{T\times (H//16) \times (W//16) \times C}$ from temporal memory to read out the feature corresponding to the target feature.
By combining the similarity matrix with the memorized ID values $\mathbf{v}\in \mathbb{R}^{T\times (H//16) \times (W//16) \times C}$, we compute the readout features $F_R$ related to the object targets
and predict the final target masks using a decoder.
%
% The mask decoder fuses object-enhanced and multi-scale features from the backbone to predict the final mask.
%
The overall process on the $t$-th frame can be formulated as:
% \begin{equation}
% \begin{aligned}
%  M_t=&\phi\left(P_{m e m}\left(\operatorname{M H A_{T r i}}\left(\mathbf{F}_t, \mathbf{F}_r, \mathbf{F}_{i d}\right), \textbf{M}_{t a r g e t}\right),\right. \\
% & \left.\mathbf{F}_t^{1 / 8}, \mathbf{F}_t^{1 / 4}\right), 
% \end{aligned}
% \end{equation}
\begin{equation}
\begin{aligned}
 M_t = \phi\Bigl(P_{mem}\bigl(\operatorname{STML}(\mathbf{F}_t, \mathbf{F}_r, \mathbf{F}_{o}), \textbf{M}_{target}\bigr), \mathbf{F}_t^{1/8}, \mathbf{F}_t^{1/4}\Bigr), 
\end{aligned}
\end{equation}
% where $\operatorname{MHA_{STML}}$ denotes the process of the Spatial-temporal Multi-level association model, 
where $\textbf{M}_{target}$ represents the memorized features, 
$P_{mem}$ indicates the function of the ID association module,
$\mathbf{F}_t^{1/4}$ and $\mathbf{F}_t^{1/8}$ are the multi-scale features of frame $t$, and $\phi$ is the decoder function, which predicts the target masks.

\subsection{Spatial-Temporal Multi-Level Feature Association}
\label{splitattention}
\begin{figure*}
% \vspace{-2mm}
  \centering
   %\fbox{\rule{0pt}{2in} \rule{0.9\linewidth}{0pt}}
   \includegraphics[width=.9\linewidth]{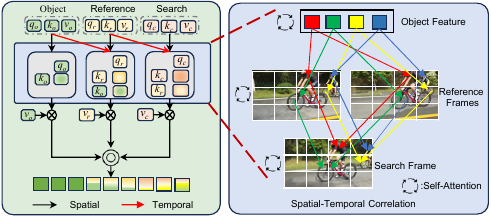}
   \caption{\textbf{Illustration of the proposed spatial-temporal correlation.} Given two reference frames as examples. The object features conduct self-attention and the reference features perform attention both with themselves and object features. The target feature undergoes attention with both itself and the reference feature simultaneously. }
   \label{fig:correlation}
   \vspace{-3mm}
\end{figure*}

% The previous works of VOS separated feature extraction from relation matching, which posed challenges in grasping finer target representations and managing extensive pre-training. 
The spatial-temporal multi-level feature association model is proposed to associate the target frame, reference frames from the spatial memory, and object features to learn comprehensive target-aware features for the subsequent target differentiation and mask prediction.
%
% We propose a module for tri-flow association, designed for feature modeling and propagation, that effectively integrates the test frame, reference frames, and object features within a single block.
%
As shown in Figure~\ref{fig:correlation}, the main component of the STML module is the spatial-temporal correlation block, which uses the Multi-Head Attention block~\cite{attention} to facilitate the interaction between different types of input information.
% As we have three types of inputs including test embeddings, reference embeddings, and object embeddings, the most important is how to model the information propagation among them.
%First, the attention calculation can be written as:
% The  with the tri-attention operation can be formulated as:
% \begin{equation}
% \begin{aligned}
% \mathbf{F}_t, \mathbf{F}_r, \mathbf{F}_{i d}=\operatorname{MHA_{Tri}}\left(\mathbf{F}_t, \mathbf{F}_r,\mathbf{F}_{i d}\right).
% \end{aligned}
% \end{equation}
As directly computing the cross-attention between all three kinds of input features is inefficient, we decompose the calculation into the operations of object self-attention $\textbf{A}^{o}_{attn}$, reference object enhancement $\textbf{A}^{r}_{attn}$, and test reference correlation $\textbf{A}^{t}_{attn}$.

%We show the calculation of tri-attention in detail.
For object features, we conduct self-attention to enhance the distinction between different targets. The self-attention of the object features $\textbf{A}^{o}_{attn}$ is computed as:
\begin{equation}
\begin{aligned}
\textbf{A}^{o}_{attn} & =  \textit{f}_{softmax}\left(\frac{q^{o} {k^{o}}^\top}{\sqrt{d}}\right) v^{o},
\end{aligned}
\end{equation}
%Three input tokens all perform self-attention within themselves.
where $\textit{f}_{softmax}$ represents the operation of SoftMax,
$q^{o}$, $k^{o}$, and $v^{o}$ are generated by projecting $\mathbf{F}_{o}$.
We then use the object features to enhance each reference feature, which is performed by correlating the object features with every reference feature.
For each reference frame feature $i$, the asymmetric attention is defined as:
\begin{equation}
\begin{aligned}
k^{r_{i}}_{tem} & = \textit{f}_{concat}\left(W^{K}\left[\mathbf{F}_{r}^{i}\right], W^{K}\left[\mathbf{F}_{o}\right]\right), \\
v^{r_{i}}_{tem} & = \textit{f}_{concate}\left(W^{V}\left[\mathbf{F}_{r}^{i}\right], W^{V}\left[\mathbf{F}_{o}\right]\right), \\
\textbf{A}^{r_{i}}_{attn} & =  \textit{f}_{softmax}\left(\frac{q^{r_{i}} {k^{r_{i}}_{tem}}^\top}{\sqrt{d}}\right) v^{r_{i}}_{tem},
\end{aligned}
\end{equation}
where $*_{tem}$ is a temporary features and $\textit{f}_{concat}$ is the operation to concatenate features. 
Our rationale for individually segmenting each reference frame feature during information correlation is to ensure the reference features remain uninfluenced by one another, preserving their discrete integrity.

Then, the asymmetric information propagation is performed between the feature of the target and references.
\begin{equation}
\begin{aligned}
k^{t}_{tem} & = \textit{f}_{concat}\left(W^{K}\left[\mathbf{F}_{t}\right], W^{K}\left[\mathbf{F}_{r}\right]\right), \\
 v^{t}_{tem} & = \textit{f}_{concat}\left(W^{V}\left[\mathbf{F}_{t}\right], W^{V}\left[\mathbf{F}_{r}\right]\right), \\
\textbf{A}^{t}_{attn} & = \textit{f}_{softmax}\left(\frac{q^t  {k^{t}_{tem}}^\top}{\sqrt{d}}\right) v^t_{tem}.
\end{aligned}
\end{equation}

In this manner, the target feature only receives the information from the reference feature, and the references remain unchanged during this step.
With the multi-level association block, the information can also propagate to the target feature.
%
% In summary, our goal is to achieve asymmetric propagation among three distinct types of information during the information dissemination process: test embedding, reference embedding, and object embedding.

%
The self-attention among object features serves to differentiate various objects. 
Each reference frame feature interacts solely with its respective frame and the object features, ensuring no information interference between reference frames while concurrently enhancing target data. 
In addition to self-attention, the test feature engages with reference features, assimilating valuable information from them. 
Such a strategic configuration is instrumental in sharpening the target details within reference features and enriches the target information in the test frame.
The intuitive representation of STML in one block is shown in~\ref{tab:overall}, where the correlation among three types of features can be strengthened by our mechanism.
% \subsection{Temporal ID Association}  %要更体现内容
\subsection{Spatial-Temporal Memory} 
\label{memorymanagement}
In this section, we give the updation of spatial-temporal memory and how to utilize the temporal memory for object feature generation.\\
\textbf{Object Feature Generation.} 
We propose a straightforward method to generate object ID features. 
Object features $\mathbf{F}_{0}\in \mathbb{R}^{n \times C \times (H//16) \times (W//16)}$ is initialized as all zeros and then generated by interacting with the first frame and ground-truth mask by a feature enhancement layer.
Object ID features are generated by pooling the memorized ID values while processing frame $t$.
\begin{equation}
\mathbf{F}_{o}=\textit{f}_{o}\left(\mathbf{v}\right)\in \mathbb{R}^{n \times C},
\end{equation}
where $\textit{f}_{o}$ projects the memorized ID features like MaxPooling.
With the object features, we model the propagation of frame $t$ in the proposed STML module.
\textbf{Spatial Memory Update.}
After completing the target prediction, the target frame is updated into the spatial memory according to a first-in-first-out strategy. 
To ensure that targets are not lost in long-term scenarios, spatial memory will not discard the frame that first provided the target.\\
% \textbf{ID Association.}
% % Our memory adopts a similar design of XMem~\cite{xmem}, organized by working memory and long-term memory.
% % We do not claim our contributions in this part.
% The memorized features consist of two additional variables, key $\mathbf{Fk}\in \mathbb{R}^{T\times (H//16) \times (W//16) \times C}$ , from the STML module and object ID values $\mathbf{v}\in \mathbb{R}^{T\times (H//16) \times (W//16) \times C}$, generated by a light encoder after the prediction.
% The matching process is similar to the attention layer.
% % The memorized keys and values are encoded from the history frames with their masks, leading to high-level appearance matching.
% We first calculate the query-key affinity matrix to read out the feature corresponding to the test feature.
% \begin{equation}
% \mathbf{S}_{i j}=\frac{\exp \left(d\left(\mathbf{F}_t, \mathbf{Fk}_j\right)\right)}{\sum_m \exp \left(d\left(\mathbf{F}_t, \mathbf{Fk}_m\right)\right)},
% \end{equation}
% where $\mathbf{S}$ is the affinity matrix and $d(\dot, \dot)$ is the anisotropic L2 function to get the similarity between two inputs.
% %
% Finally, by combining the similarity matrix with the memorized ID values, we compute the readout features $F_R$ related to the object targets as:
% \begin{equation}
% F_R=\mathbf{S}\mathbf{v}.
% \end{equation}
% Then the feature after the ID association is processed to predict the final mask of each target.
% %
% The final output is generated by soft aggregation.
\textbf{Temporal Memory Update.} 
After obtaining the prediction output of the current frame, we employ a simple fusion strategy to add the current frame into memory. 
The target feature of STML is updated as the key into the temporal memory. 
By concatenating the output masks of different targets with the target frame and encoding the features, the ID values of different targets are obtained and stored in the temporal memory.
Similar to XMem~\cite{xmem}, when the memory reaches its specified capacity, we update the elements in memory using the least-frequently-used (LFU) eviction algorithm.
% After retrieving the readout features of each target from long-term memory, we employ a feature enhancement layer to augment the target IDs with $\mathbf{h}_{t-1}$.
% The ID propagation module receives the test feature and the previous object features $\mathbf{h}_{t-1}$ from short-term memory to generate.
% Here, we complete the ID association.
% The mask decoder then processes the enhanced features to generate the final output.\\
% new ablation -------------------------------------------
\begin{table}
\vspace{-3mm}
\centering
\caption{\textbf{Ablation study on the DAVIS 2017 val and test datasets.} It shows the performance gains of each component in terms of region similarity ($\mathcal{J}$), contour accuracy ($\mathcal{F}$), and their average score($\mathcal{J}\&\mathcal{F}$). }
\label{tab:object}
\rowcolors{2}{gray!25}{white}
\renewcommand\arraystretch{1.2}
\setlength{\tabcolsep}{2mm}{
\small   % 字体大小设置为 'small'
\resizebox{.88\linewidth}{!}{
\begin{tabular}{c|ccc|ccc|c}
\toprule
{\textbf{Dataset}}  & \multicolumn{3}{c|}{\textbf{DAVIS 2017 val}} & \multicolumn{3}{c}{\textbf{DAVIS 2017 test}} & \textbf{FPS}\\
{\textbf{Variant}} & $\mathcal{J}\&\mathcal{F}$ & $\mathcal{J}$ & $\mathcal{F}$ & $\mathcal{J}\&\mathcal{F}$ & $\mathcal{J}$ & $\mathcal{F}$              \\
\midrule
W/O Obj & 88.2& 84.8&91.6 &83.8 &80.1 &86.8 & 11.2\\
W/O Spatial Memory   &87.8 &84.8 &90.9 &83.5 &80.4&86.6  &22.1          \\
W/O Temporal Memory &88.1 &84.9 &90.9 &81.0 &77.5 &84.6 &3.5 \\
Joint Self-Attn &87.8 &84.8 &90.8 &82.7 &79.5 &85.8 & 9.6 \\
Full version (Proposed method)      & \textbf{88.9} & \textbf{86.0} & \textbf{91.8} &\textbf{85.6} &\textbf{81.9} &\textbf{89.2} & 10.7\\
\bottomrule
\end{tabular}
}
}
\vspace{-6mm}
\end{table}

\subsection{Training and Inference}
\textbf{Training.}
Our method follows the three-step training scheme used by previous methods~\cite{stcn, vos_corr,xmem}.
We first conduct the synthetic static pre-training on various static datasets~\cite{CSSD,fss1000,SOD,cascadepsp,learning}.
Then we fine-tune the model on video datasets, including DAVIS~\cite{davis2017} and YouTube-VOS~\cite{ytvos2018}.
To improve the performance, we also perform training on the BLK30k dataset~\cite{modular}.
We use the strategy of curriculum sampling, and the sampled sequence length is set to $5$.
Two of the past frames in spatial memory are randomly selected as the reference frames.
The training loss uses a combination of bootstrapped cross entropy loss and Dice loss, with equal weights of 0.5 for each.
\begin{table}
\vspace{-3mm}
\centering
% \setlength{\extrarowheight}{0pt}
% \addtolength{\extrarowheight}{\aboverulesep}
% \addtolength{\extrarowheight}{\belowrulesep}
% \setlength{\aboverulesep}{0pt}
% \setlength{\belowrulesep}{0pt}
\caption{\textbf{Comparison over different backbone and pre-trained weights.} It shows that our method achieves the best performance under the same backbone and pre-trained weights.}
\renewcommand\arraystretch{1.2}
\setlength{\tabcolsep}{2mm}{
\label{tab:conv_comparison}
\resizebox{.85\textwidth}{!}{
\begin{tabular}{l|l|l|l|l|l|l} 
\toprule
\textbf{Methods}                                                       & \textbf{Backbone} & \textbf{Pretrained} & \textbf{D17 val} & \textbf{D17 test} & \textbf{Y19 val} & \textbf{FPS}  \\ 
\midrule
XMem~\cite{xmem}                                                                   & ConvViTb          & MAE                 & 85.7             & 82.1              & 84.7             & 22.3          \\
\rowcolor[rgb]{0.875,0.875,0.875} {\cellcolor{white}}DeAOTL~\cite{deaot}                               & ConvViTb          & MAE                 & 84.2             & 80.5              & 83.6             & 8.9           \\
SimVOS-BS~\cite{SimVOS}                                                              & ViTb              & MAE                 & 87.1             & 82.2              & 79.3             & 7.4           \\
\rowcolor[rgb]{0.875,0.875,0.875} {\cellcolor{white}} & ViTb              & MOCOV3              & 81.3             & -                 & -                & 3.1           \\
\multirow{-2}{*}{SimVOS~\cite{SimVOS}} & ViTb              & MAE                 & 88.0             & 80.4              & 84.2             & 3.1           \\ 
\hline
{\cellcolor{white}} & ViTb              & MOCOV3              & 84.1             & 79.7              & 83.4             & 11.0          \\
\rowcolor[rgb]{0.875,0.875,0.875} {\cellcolor{white}}                                   & ViTb              & MAE                 & \textbf{88.2}    & \textbf{82.1 }    & \textbf{85.0}    & 11.0          \\
{\cellcolor{white}} & ConvViTb          & Scratch             & 83.6             & 77.0              & 82.3             & 10.7          \\
\rowcolor[rgb]{0.875,0.875,0.875} \multirow{-4}{*}{{\cellcolor{white}}Ours}             & ConvViTb          & MAE                 & \textbf{88.9 }   & \textbf{85.6 }    & \textbf{86.3 }   & 10.7          \\
\bottomrule
\end{tabular}
}
}
\vspace{-6mm}
\end{table}
\vspace{2mm}
\noindent\textbf{Inference.} 
We use a first-in-first-out queue to memorize the reference frames in spatial memory for both DAVIS dataset~\cite{davis2017} and YouTube-VOS dataset~\cite{ytvos2018}.
Specifically, we keep the first frame unchanged in our memory.
For the DAVIS dataset~\cite{davis2017}, the frame is added to the queue every 3rd frame, while the 5th frame is for the YouTube-VOS dataset.
All the results are generated in the NVIDIA V100 GPU.
% For the inference on DAVIS~\cite{davis2017} we use the default 480p 24FPS videos and 6FPS on YouTube-VOS dataset~\cite{ytvos2018}.
\section{Experiments}
In this section, we conduct experiments to evaluate the proposed method on numerous VOS datasets from multiple aspects.
We first introduce the implementation details of our model.
Then, we conduct ablation studies to validate the effectiveness of the spatial-temporal memory and STML module.
Finally, we evaluate the overall performance of the proposed method against state-of-the-art approaches using both qualitative and quantitative measures.
%
% The is analyzed in Table~\ref{tab:overall} compared with the state-of-the-art methods.
% %
% Visualized results on challenging sequences are also provided for a comprehensive qualitative analysis.
%
More detailed results and implementation details are given in the supplemental materials.
\subsection{Implementation Details}
We use the ViT with different pre-trained weights to construct the STML module. 
The decoder consists of different residual upsampling blocks, which gradually fuse the multi-scale features to predict the final single-channel logits.
We use AdamW~\cite{adam} with learning rate $3e^{-5}$ and a weight decay of 0.05 as the optimizer.
The pretrain iteration on the static dataset is 150k with batch size 16, and stage 1 is performed with 250k iterations.
The main training stage on DAVIS and YouTube-VOS is conducted with 160k iterations with batch size 16.
The learning rate reduces by 10 times after 100k iterations.
More experimental settings can be found in the supplement material.
%
% related works:
% 1、temporal
% 2、spatial
% 3、spatial-temporal
% --------------------------------------------------------------------------------------
%--sota comparison------------------------------------
\begin{table*}
  \centering
  \caption{\textbf{Quantitative comparisons on the DAVIS 2017, YouTube-VOS 2018 \& 2019 dataset.} The best two results are shown in \textcolor[rgb]{1,0,0}{red} and \textcolor[rgb]{0,0,1}{blue} color. In the table, * and † denote the models are pre-trained using the additional static image datasets and the large BL30K dataset, respectively.}
  \rowcolors{2}{gray!25}{white}
  \small
  \renewcommand\arraystretch{1.3}
  \resizebox{1.0\linewidth}{!}{
  \begin{tabular}{l|ccc|ccc|ccc|ccccc|ccccc}
    \toprule
    {\textbf{Dataset}} & \multicolumn{3}{c|}{\textbf{DAVIS 2016 val}} & \multicolumn{3}{c|}{\textbf{DAVIS 2017 val}} & \multicolumn{3}{c|}{\textbf{DAVIS 2017 test}} & \multicolumn{5}{c|}{\textbf{YouTube-VOS 2018 val}} & \multicolumn{5}{c}{\textbf{YouTube-VOS 2019 val}}\\
{\textbf{Method}} & $\mathcal{J}\&\mathcal{F}$ & $\mathcal{J}$ & $\mathcal{F}$ & $\mathcal{J}\&\mathcal{F}$ & $\mathcal{J}$ & $\mathcal{F}$ & $\mathcal{J}\&\mathcal{F}$ & $\mathcal{J}$ & $\mathcal{F}$  & $\mathcal{G}$ & $\mathcal{J}_s$ & $\mathcal{F}_s$ & $\mathcal{J}_u$ & $\mathcal{F}_u$ & $\mathcal{G}$ & $\mathcal{J}_s$ & $\mathcal{F}_s$ & $\mathcal{J}_u$ & $\mathcal{F}_u$\\
    \midrule
    
    SST~\cite{sstvos} & - &- &- &82.5 &79.9  &85.1   &- &- &- &81.7& 81.2 &- &76.0 &- &81.8 &80.9 &- &76.6 &-\\
    
    JOINT~\cite{joint}  & - &- &- & 83.5 &80.8 &86.2 &- &- &- &83.1 &81.5 &85.9 &78.7 &86.5 &82.8 &80.8 &84.8 &79.0 &86.6\\
    
    XMem~\cite{xmem}   & - &- &- & 84.5 & 81.4 & 87.6 & 79.8 & 76.3 & 83.4 & 84.3 & 83.9 & 88.8 & 77.7 & 86.7 & \textcolor{blue}{84.2} & \textcolor{blue}{83.8} & \textcolor{blue}{88.3} & 78.1 & \textcolor{blue}{86.7}\\
    VOS$-$Corr~\cite{vos_corr}  &-&-&-&86.1 &82.7 &89.5 &\textcolor{blue}{81.0} &\textcolor{blue}{77.3} &\textcolor{blue}{84.7} &\textcolor{blue}{85.6} &\textcolor{blue}{84.9} &\textcolor{blue}{89.7} &\textcolor{blue}{79.0} &\textcolor{blue}{87.8} &-&-&-&-&-\\
    SimVOS-B~\cite{SimVOS} & \textcolor{red}{92.9} & \textcolor{red}{91.3} & \textcolor{red}{94.4}  & \textcolor{blue}{88.0} & \textcolor{blue}{85.0} & \textcolor{blue}{91.0} & 80.4 & 76.1& 84.6 & - & - & - & - & - & \textcolor{blue}{84.2} & 83.1 & 87.1& \textcolor{blue}{79.1} & 84.7\\
    %JointFormer &- &- &- &\textcolor{red}{89.1} &85.9 &\textcolor{red}{92.2} &\textcolor{red}{87.0} &\textcolor{red}{83.4} &\textcolor{red}{90.6} &\textcolor{red}{86.0} &86.0 &\textcolor{red}{91.0} &79.5 &87.5 &86.2 &\textcolor{red}{85.7} &\textcolor{red}{90.5} &80.4 &88.2 \\
    % Add all other rows in a similar fashion
    \textbf{STMA} & 91.6 & 90.3 & 93.0  & \textcolor{red}{\textbf{88.9}} &\textcolor{red}{\textbf{86.0}} &\textcolor{red}{\textbf{91.8}} &\textcolor{red}{\textbf{85.6}} &\textcolor{red}{\textbf{81.9}} &\textcolor{red}{\textbf{89.2}} & \textcolor{red}{{86.1}}   & \textcolor{red}{\textbf{85.1}}   & \textcolor{red}{\textbf{90.1}} & \textcolor{red}{\textbf{80.1}}   & \textcolor{red}{\textbf{89.1}}&\textcolor{red}{\textbf{86.3}} &\textcolor{red}{\textbf{85.0}} &\textcolor{red}{\textbf{89.5}} &\textcolor{red}{\textbf{81.3}} &\textcolor{red}{\textbf{89.2}} \\

    \hline
    STM~\cite{STM} * & 89.3 &88.7 &89.9 & 81.8 & 79.2 & {84.3} & {-}   & {-}   & {-}   & 79.4 & 79.7 & 84.2 &72.8 &80.9& {-}   & {-}   & {-} & {-}   & {-}\\
    AFB-URR~\cite{afb-urr} * & - &- &- & 76.9 & 74.4 & {79.3} & {-}   & {-}   & {-}   & 79.6 & 78.8 & 83.1  &74.1 &82.6 & {-}   & {-}   & {-} & {-}   & {-}\\
    CFBI~\cite{CFBI} *    & - &- &- & 81.9 & 79.1 & 84.6   & 74.8  & 71.1  & 78.5  & 81.4 &81.1  &85.8  &75.3   & 83.4  & 81.0 & 80.6 &85.1 & 75.2 & 83\\
    MiVOS~\cite{modular}* & - &- &- & 84.5 &81.7 &87.4 &78.6 &74.9 &82.2 &82.6 &81.1 &85.6 &77.7 &86.2 &82.4 &80.6 &84.7 &78.1 &86.4\\
    STCN~\cite{stcn} * & 91.6 &90.8 & 92.5 & 85.3 &82.0 &88.6 &77.8 &74.3 &81.3 &84.3 &83.2 &87.9 &79.0 &87.3 &84.2 &82.6 &87.0 &79.4 &87.7  \\
    Swin-B-AOT-L~\cite{aot} * &92.0 &90.7 &93.3 & 85.4 &82.4 &88.4 &81.2 &77.3 &85.1 &85.1 &85.1 &90.1 &78.4 &86.9 &85.3 &84.6 &89.5 &79.3 &87.7   \\
    SwinB-DeAOT-L\cite{deaot} * &92.9 &91.1 &\textcolor{blue}{94.7} & 86.2 &83.1 &89.2 &82.8 & 78.9 &\textcolor{blue}{86.7} &86.3 &\textcolor{blue}{85.4} &\textcolor{red}{90.7} &80.1 &89.0 &86.4 &\textcolor{blue}{85.4} &\textcolor{red}{90.3} &80.5 &89.3 \\ 
    XMem\cite{xmem} * & 91.5 &90.4 &92.7 & 86.2 &82.9 &89.5 &81.0 &77.4 &84.5 &85.7 &84.6 &89.3 &80.2 &88.7 &85.5 &84.3 &88.6 &80.3 &88.6 \\
    VOS-Corr~\cite{vos_corr} &92.2 &91.1 &93.3 &87.7 &\textcolor{blue}{84.1} &\textcolor{blue}{91.2} &82.0 &78.3 &85.6 &\textcolor{red}{86.9} &\textcolor{red}{85.5} &\textcolor{blue}{90.2} &\textcolor{red}{81.6} &\textcolor{red}{90.4} &\textcolor{blue}{86.6} &85.3 &89.8 &\textcolor{blue}{81.4} &\textcolor{red}{89.8} \\
    ISVOS~\cite{ISVOS} * &\textcolor{blue}{92.6} &\textcolor{blue}{91.5} &93.7& \textcolor{blue}{87.1} &83.7 &90.5 &\textcolor{blue}{82.8} &\textcolor{blue}{79.3} &86.2 &86.3 &85.5 &90.2 &80.5 &88.8 &86.1 &85.2 &89.7 &80.7 &88.9\\
    %JointFormer* &92.1 &90.6 &93.6 &89.7 &86.7 &92.7 &\textcolor{red}{87.6} &\textcolor{red}{84.2} &\textcolor{red}{91.1} &\textcolor{red}{87.0} &\textcolor{red}{86.2} &\textcolor{red}{91.0} &\textcolor{red}{81.4} &\textcolor{red}{89.3} &\textcolor{red}{87.0} &\textcolor{red}{86.1} &\textcolor{red}{90.6} &\textcolor{red}{82.0} &89.5 \\

    \textbf{STMA} * & \textcolor{red}{\textbf{93.4}} & \textcolor{red}{\textbf{91.9}} & \textcolor{red}{\textbf{94.8}} &  \textcolor{red}{\textbf{90.4}} &\textcolor{red}{\textbf{87.3}} &\textcolor{red}{\textbf{93.5}} &\textcolor{red}{\textbf{87.1}} &\textcolor{red}{\textbf{83.7}} &\textcolor{red}{\textbf{90.5}}& \textcolor{blue}{86.4}   & {85.0}   & {89.6} & \textcolor{blue}{81.5}   & \textcolor{blue}{89.3} &\textcolor{red}{\textbf{86.8}} &\textcolor{red}{\textbf{85.6}} &\textcolor{blue}{90.0} &\textcolor{red}{\textbf{81.8}} &\textcolor{blue}{89.5} \\
    \hline
    XMem~\cite{xmem}† &  92.0 &90.7 &93.2 & 87.7 &84.0 &91.4 &81.2 &77.6 &84.7 &86.1 &85.1 &89.8 &80.3 &\textcolor{blue}{89.2} &85.8 &84.8 &89.2 &80.3 &88.8\\
    ISVOS\cite{ISVOS} † &\textcolor{blue}{92.8} &\textcolor{blue}{91.8} &\textcolor{blue}{93.8}& \textcolor{blue}{88.2} & \textcolor{blue}{84.5} & \textcolor{blue}{91.9} & \textcolor{blue}{84.0} & \textcolor{blue}{80.1} & \textcolor{blue}{87.8} & \textcolor{blue}{86.7} & \textcolor{red}{86.1} & \textcolor{red}{90.8} & \textcolor{blue}{81.0} & 89.0 & \textcolor{blue}{86.3} & \textcolor{blue}{85.2} & \textcolor{blue}{89.7} & \textcolor{blue}{81.0} & \textcolor{blue}{89.1}\\
    %JointFormer † &92.4 &90.4 &94.4 &90.1 &87.0 &93.2 &\textcolor{red}{88.1} &\textcolor{red}{84.7} &\textcolor{red}{91.6} &\textcolor{red}{87.6} &\textcolor{red}{86.4} &\textcolor{red}{91.0} &\textcolor{red}{82.2} &\textcolor{red}{90.7} &\textcolor{red}{87.4} &\textcolor{red}{86.5} &\textcolor{red}{90.9} &82.0 &\textcolor{red}{90.3} \\
    \textbf{STMA} † &\textcolor{red}{\textbf{93.5}} & \textcolor{red}{\textbf{92.1}} & \textcolor{red}{\textbf{94.9}} &\textcolor{red}{\textbf{90.8}} &\textcolor{red}{\textbf{87.8}} &\textcolor{red}{\textbf{93.9}}& \textcolor{red}{\textbf{87.7}} &\textcolor{red}{\textbf{84.4}} &\textcolor{red}{\textbf{91.0}} & \textcolor{red}{\textbf{87.0}}   & \textcolor{blue}{85.7}   & \textcolor{blue}{90.2} & \textcolor{red}{\textbf{82.0}}   & \textcolor{red}{\textbf{90.0}}&\textcolor{red}{\textbf{87.1}} &\textcolor{red}{\textbf{85.6}} &\textcolor{red}{\textbf{90.0}} &\textcolor{red}{\textbf{82.4}} &\textcolor{red}{\textbf{90.1}} \\
    \bottomrule
  \end{tabular}}
  \label{tab:overall}
  \vspace{-3mm}
\end{table*}
\subsection{Ablation Study}
In this section, we conduct experiments on different datasets to validate the effectiveness of our proposed approach.
In Table~\ref{tab:object}, We train five different variants of our method on the YouTube-VOS and DAVIS datasets with ConvVitb as the STML module.
In Table~\ref{tab:conv_comparison}, performance with different backbone and pre-trained weights is compared.\\
\textbf{W/O Obj}, which removes the object features from the STML module.
The reference features conduct self-attention within themselves.
The target feature employs self-attention and cross-attention with reference features.\\
\textbf{W/O Spatial Memory}, which removes the reference features from the input of the STML module.
We conduct cross-attention between target features and object features.\\
\textbf{W/O Temporal Memory}, which removes the temporal memory and ID association module.
The ID classification of different targets is similar to SimVOS, which adds the mask of every target to the corresponding frames.
We concate the mask of each target to the corresponding frames and follow the SimVOS to perform the VOS task.\\
\textbf{Joint Self-Attn} conducts a straightforward self-attention by concatenating all three inputs, object features, reference features, and target features.\\
\noindent\textbf{Effect of object features.} 
Without object features, the performance on three different validation sets all decreases.
Specifically, using object features gains 0.7\% and 2.5\% in $\mathcal{J}\&\mathcal{F}$ on the DAVIS 2017 val and DAVIS 2017 test, respectively.
The results validate the benefits of enhancing the target representation in reference and test frames by utilizing object features in the STML module.\\
% \vspace{1mm}
% \noindent\textbf{Effect of memory.} 
% In our model, short-term memory serves dual critical functions: it both enhances the features of the targets and contributes to the creation of object features. 
% %
% As shown in Table~\ref{tab:object}, omitting this component leads to a notable decrease in performance, evidenced by a 2.9\% drop in $\mathcal{J}\&\mathcal{F}$ on the DAVIS 2017 val set and a 2.0\% drop on the DAVIS 2017 test set. 
% %
% The significant reduction in all metrics validates the critical role of short-term memory in effectively modeling and propagating object features.
\noindent\textbf{Effect of the STML module.}
We present the performance of different associations among object, reference, and test features.
First, we try to remove the object features or reference features.
Both decrease the performance on the DAVIS dataset, indicating that our SPM module achieves sufficient feature correlation and target association.
Then, we explore a straightforward way to jointly perform self-attention by concatenating the three features, which gains no improvement on both datasets and even gets lower $\mathcal{J}\&\mathcal{F}$ in the test set.
Although this way generates sufficient correlation, it utilizes extra background noises than our proposed method.
The proposed STML module surpasses other information interaction methods on the DAVIS dataset.
Especially on the DAVIS 2017 test set, our method gains 2.0$+$\% in $\mathcal{J}\&\mathcal{F}$.
These comparisons demonstrate the efficiency of STML in effectively managing the association of target features, thereby making it easier for the model to distinguish each objects.\\
\noindent\textbf{Effect of backbone and pre-trained weights.} 
Table~\ref{tab:conv_comparison} shows the comparison of our methods with other SOTA methods on different backbones and pre-trained weights.
We replace the backbone of XMem and DeAOT with ConvViTb and train them on video datasets.
The results show that our method outperforms other methods substantially with the same backbone.
We also confirm the importance of pre-trained weights.
Compared to methods using transformers for ID association, our method achieved favorable inference speeds, especially when compared with SimVOS.
\begin{table}
\vspace{-3mm}
\centering
\caption{\textbf{Long-term performance on the LVOS test set. } 
The results are obtained with or without fine-tuning the training set of the LVOS dataset.
It shows that our approach performs well on long-term videos.
}
%\rowcolors{2}{white}{gray!25}
\label{tab:lvos}
\renewcommand\arraystretch{1.2}
\newcommand{\tabincell}[2]{\begin{tabular}{@{}#1@{}}#2\end{tabular}}
\setlength{\tabcolsep}{2mm}{
%\small   % 字体大小设置为 'small'
\resizebox{1\linewidth}{!}{
\begin{tabular}{cc|cccccccccc}
\toprule
\multicolumn{2}{c|}{Method}                                                       & \tabincell{c}{AFB-\\URR~\cite{afb-urr} } &\tabincell{c}{CFBI\\~\cite{CFBI}} &\tabincell{c}{STCN\\ \cite{stcn}} &  \tabincell{c}{RDE\\\cite{recurrent}}  &  \tabincell{c}{XMem\\\cite{xmem}} &  \tabincell{c}{LWL\\\cite{lwl}}  &  \tabincell{c}{AOT-\\L~\cite{aot}} &  \tabincell{c}{AOT-\\B~\cite{aot}} &  \tabincell{c}{DDMem\\\cite{lvos}} & \textbf{Ours}  \\
\midrule
% \multirow{3}{*}{\begin{tabular}[c]{@{}c@{}}Without\\Finetuning\end{tabular}} 
\multirow{3}{*}{\tabincell{c}{Without\\Finetuning} }
& $\mathcal{J}\&\mathcal{F}$ & 39.9  & 44.8 & 45.8 & 49.0  & 49.5 & 50.7 & 54.1  & 54.4  & 55.0  & \textbf{56.8}  \\
 & $\mathcal{J}$  & {\cellcolor[rgb]{0.875,0.875,0.875}} 36.2  & {\cellcolor[rgb]{0.875,0.875,0.875}} 40.5 & {\cellcolor[rgb]{0.875,0.875,0.875}} 41.6 & {\cellcolor[rgb]{0.875,0.875,0.875}} 44.4 & {\cellcolor[rgb]{0.875,0.875,0.875}} 45.2 & {\cellcolor[rgb]{0.875,0.875,0.875}} 46.5 & {\cellcolor[rgb]{0.875,0.875,0.875}} 48.7  & {\cellcolor[rgb]{0.875,0.875,0.875}} 49.3  & {\cellcolor[rgb]{0.875,0.875,0.875}} 49.9   & {\cellcolor[rgb]{0.875,0.875,0.875}} \textbf{52.7}  \\
& $\mathcal{F}$  & 43.6  & 49.0 & 50.0 & 53.5 & 53.7 & 54.8 & 59.5  & 59.4  & 60.2   & \textbf{60.9}  \\
\midrule
\multirow{3}{*}{Finetuning}  
& $\mathcal{J}\&\mathcal{F}$ & {\cellcolor[rgb]{0.875,0.875,0.875}} 40.8  & {\cellcolor[rgb]{0.875,0.875,0.875}} 44.8 & {\cellcolor[rgb]{0.875,0.875,0.875}} 48.3 & {\cellcolor[rgb]{0.875,0.875,0.875}} 50.2 & {\cellcolor[rgb]{0.875,0.875,0.875}} 50.9 & {\cellcolor[rgb]{0.875,0.875,0.875}} 50.8 & {\cellcolor[rgb]{0.875,0.875,0.875}} 54.7  & {\cellcolor[rgb]{0.875,0.875,0.875}} 54.5  & {\cellcolor[rgb]{0.875,0.875,0.875}} 55.7   & {\cellcolor[rgb]{0.875,0.875,0.875}} \textbf{57.4}  \\
 & $\mathcal{J}$  & 37.5  & 40.2 & 44.0 & 45.7 & 46.5 & 46.4 & 49.2  & 49.2  & 50.3   & \textbf{53.3}  \\
 & $\mathcal{F}$  & {\cellcolor[rgb]{0.875,0.875,0.875}} 44.1  & {\cellcolor[rgb]{0.875,0.875,0.875}} 49.4 & {\cellcolor[rgb]{0.875,0.875,0.875}} 52.5 & {\cellcolor[rgb]{0.875,0.875,0.875}} 54.6 & {\cellcolor[rgb]{0.875,0.875,0.875}} 55.3 & {\cellcolor[rgb]{0.875,0.875,0.875}} 55.2 & {\cellcolor[rgb]{0.875,0.875,0.875}} 60.2  & {\cellcolor[rgb]{0.875,0.875,0.875}} 59.8  & {\cellcolor[rgb]{0.875,0.875,0.875}} 61.2   & {\cellcolor[rgb]{0.875,0.875,0.875}} \textbf{61.5} \\
\bottomrule
\end{tabular}
}
}
\vspace{-4mm}
\end{table}
%---------------------------------------------------------------------------------------------------------------------
\subsection{State-of-the-Art Comparison}
We evaluate the proposed method by comparing it with the state-of-the-art methods from both quantitative and qualitative analyses.
All the models are trained with ConvViTb as the STML module.
The experiments are conducted on five datasets, including DAVIS2016 val set, DAVIS 2017 val \& test set, and YouTube-VOS 2018 \& 2019 val set.
%,
We discuss the detailed results on every dataset below.
\begin{figure*}
  \centering
   %\fbox{\rule{0pt}{2in} \rule{0.9\linewidth}{0pt}}
   \includegraphics[width=.98\linewidth]{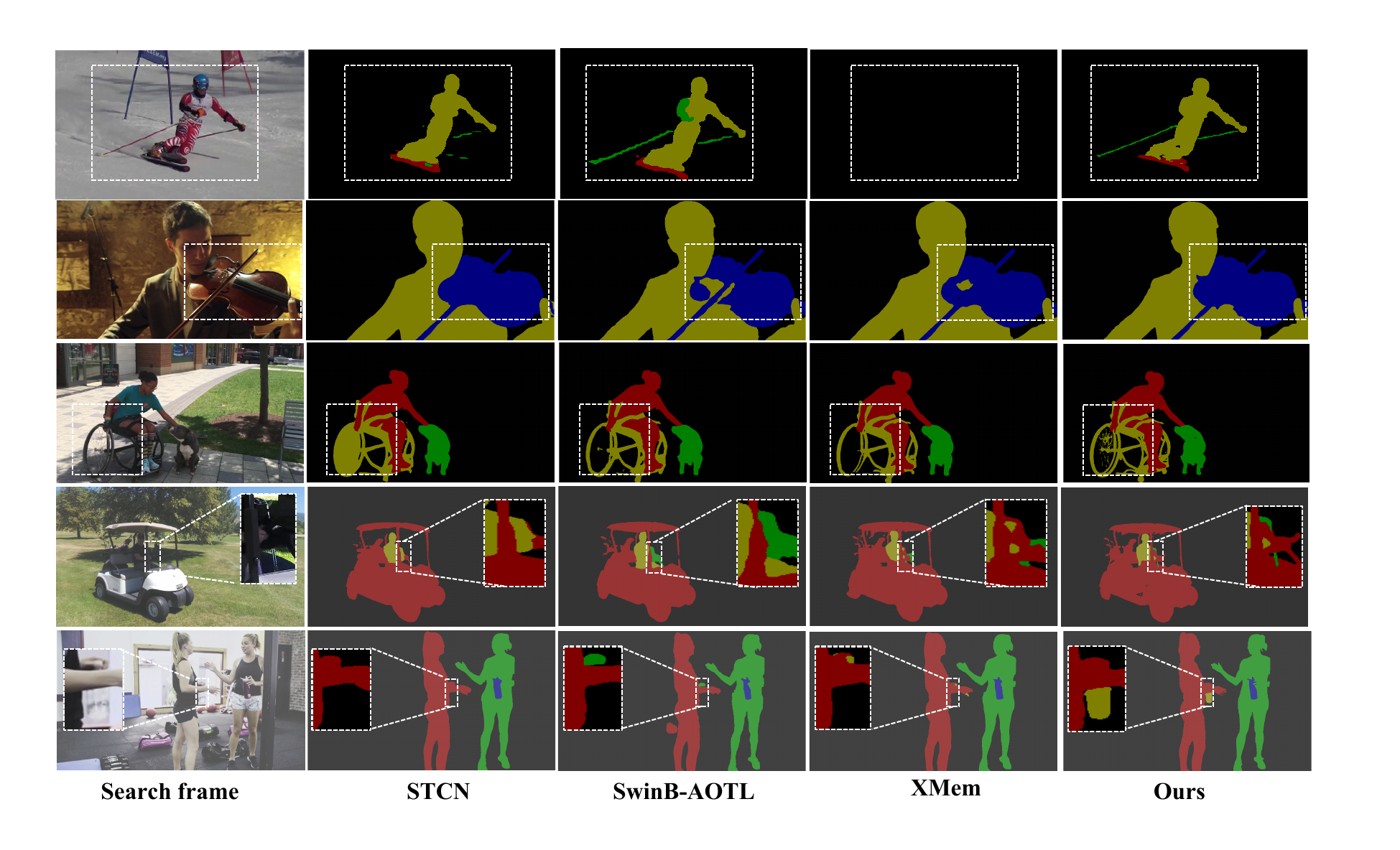}
   \caption{\textbf{Visualized results on sequences with small and faint objects}. It shows that our method generates finer masks compared to the state-of-the-art methods.}
   \label{fig:case}
   \vspace{-4mm}
\end{figure*}

% \vspace{2mm}
\noindent\textbf{Quantitative Analysis.}\\
Table~\ref{tab:overall} compares our method with previous state-of-the-art models on five benchmarks.

\noindent\textbf{DAVIS 2016}~\cite{davis2017} is a classical VOS dataset that offers 50 high-quality, single-object video sequences with frame-by-frame pixel-level annotations for developing and validating video segmentation algorithms.
After pertaining, our method gains significant improvement (93.5\% $\mathcal{J}\&\mathcal{F}$) on the val set.
Extensive information enhancement of targets in the STML module makes distinguishing and identifying target objects easier during model prediction.\\
\textbf{DAVIS 2017}~\cite{davis2017}, an extension of DAVIS2016,  is a benchmark that offers densely annotated, high-quality, full-resolution videos with multiple objects of interest.
The val set has 30 videos, and the test set has more videos in challenging scenarios.
Table~\ref{tab:overall} shows our method achieves gains better performance in both DAVIS 2017~\cite{davis2017} val and test-dev set.
Even without pre-training on static images, our model still shows favorable results, 88.9\% on val and 85.6\% on test-dev.
%
% Our model experiences further enhancements in performance through pre-training at various stages.
%
The improvements validate the efficacy of our model in multi-object and small-target scenarios.\\
\textbf{YouTube-VOS 2018}~\cite{ytvos2018} has 3471 videos with 65 categories for training and 474 videos for validation.
In the validation, there are 26 categories that the model has not seen in its training, enabling us to evaluate its generalization ability for class-agnostic targets.
Table~\ref{tab:overall} illustrates our method achieves enhanced performance compared to previous state-of-the-art models.
\begin{figure*}
  \centering
   %\fbox{\rule{0pt}{2in} \rule{0.9\linewidth}{0pt}}
   \includegraphics[width=1.0\linewidth]{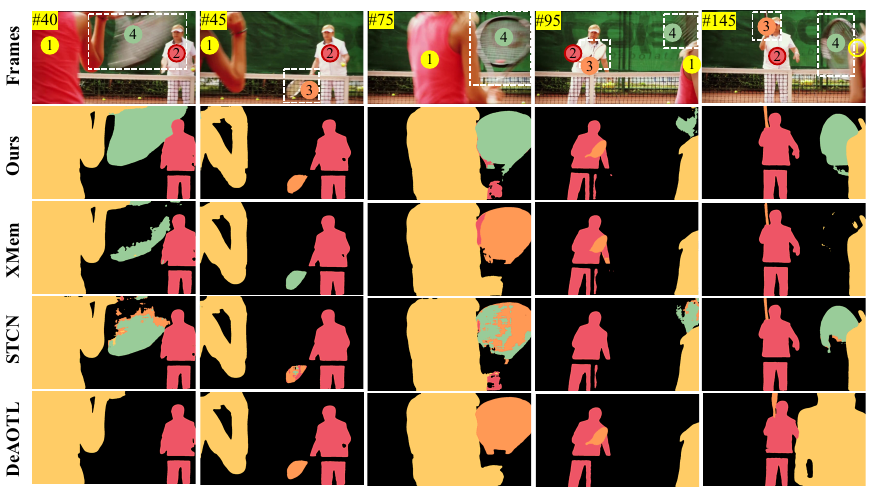}
   \caption{\textbf{Visualized results on sequences with complicated ID connections.} 
   The proposed method performs well in tracking the tennis rackets, which demonstrates excellent performance in terms of ID propagation.}
   \label{fig:idcase}
   \vspace{-4mm}
\end{figure*}
Specifically, our model performs better in unseen categories, with 82.0\% in $\mathcal{J}_{u}$ and 90.0\% in $\mathcal{F}_{u}$, indicating that our method possesses a more powerful generalization ability.\\
\textbf{YouTube-VOS 2019} is an extension of YouTube-VOS 2018, featuring a greater number of masked targets and including more challenging sequences in its val set.
In Table~\ref{tab:overall}, our method achieves competitive performance against the state-of-the-art methods.
Notably, our model gets favorable performance without any pre-training.\\
\textbf{LVOS}~\cite{lvos} is a new benchmark for evaluating video object segmentation algorithms in long-term challenging sequences with diverse scenarios and pixel-level annotations. 
We evaluate our method on the LVOS~\cite{lvos} test set without fine-tuning.
Table~\ref{tab:lvos} shows that our method achieves \textbf{56.8\%} $\mathcal{J}\&\mathcal{F}$ and performs better than previous approaches, which demonstrates the robustness of our memory bank and object features when dealing with long-term scenarios.

\vspace{2mm}
\noindent\textbf{Qualitative Analysis.}\\
We visualize some challenging sequences, including scenarios involving small and faint targets, as well as those where targets closely resemble the background.
%
% Small object segmentation
In Figure~\ref{fig:case}, our approach shows favorable segmentation performance against the state-of-the-art methods.
% we can observe that, compared with current state-of-the-art methods, our approach shows better segmentation performance.
%especially in challenging scenarios involving small and similar targets.
%
For instance, our model accurately segments targets and provides detailed results in a video that includes wheelchairs, golf clubs, and transparent bottles.
% For instance, in examples involving the wheelchair, the golf club, and the transparent bottle, our model accurately segments the targets and provides detailed results.
%
This is attributed to our proposed spatial-temporal multi-level feature association, which enables our model to capture more detailed target information by sufficient interaction, thereby facilitating precise segmentation in these challenging scenarios 
Additionally, incorporating object features in our model ensures consistent maintenance of the target ID throughout extended video sequences, such as in the case of skating.
% Objects association.
Figure~\ref{fig:idcase} displays visualized results from different methods within a video sequence.
Other methods encounter issues with ID misalignment when distinguishing between two badminton rackets.
Although SwinB-DeAOTL~\cite{aot} accurately segments target 4 at 75\textit{th} frame, the ID assignment is imprecise.
In comparison, our method distinguishes different objects with the same attributes, further validating the effectiveness of our framework.

\begin{figure}[t]
  \centering
   %\fbox{\rule{0pt}{2in} \rule{0.9\linewidth}{0pt}}
   \includegraphics[width=1.0\linewidth]{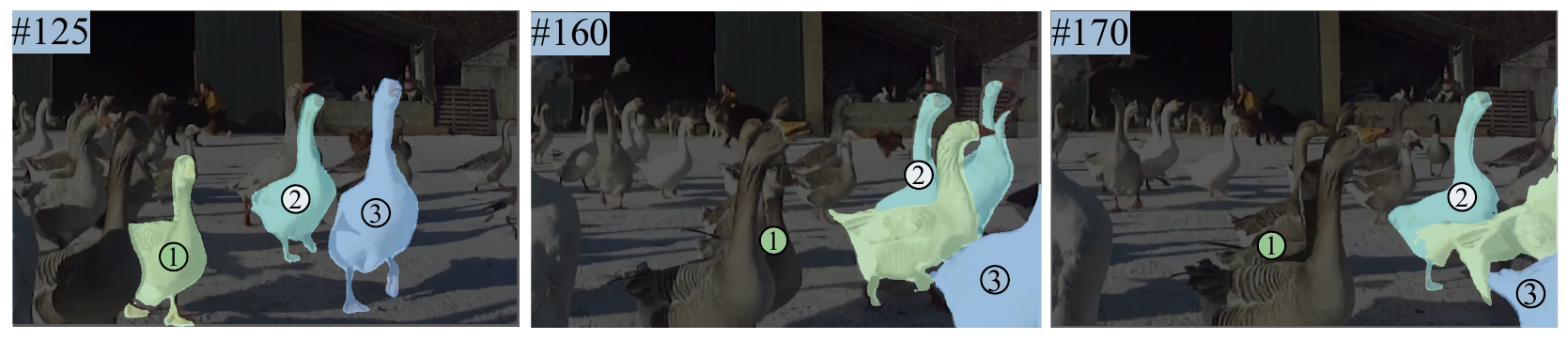}
   \caption{\textbf{Failure case.}
   %
   %We mark the targets with their ID.
   %
   It shows that our approach incorrectly associates the goose with ID 1 with a background goose due to the occlusion of the target goose and insufficient discriminative information between the geese.}
   \label{fig:limitation}
   \vspace{-4mm}
\end{figure}
\subsection{Limitations}
% 虽然我们的方法在多目标特征关联和ID传播上取得了很好的效果。
% 但是，在一些具有与目标高度相似物体和遮挡的场景中，准确进行ID的传播完成分割对我们的方法也是一个挑战。
% 如图5，我们的方法在目标1和目标2被相似目标遮挡的时候，错误的传播了ID，导致分割失败。
% 在后续的工作中，我们将着重于怎么进行准确的，去解决更复杂场景下的vos问题。
% Our method achieves favorable results in multi-target feature association and ID propagation. 
Our method enhances the segmentation precision and the accuracy of ID matching for small targets through multi-object feature correlation and ID association, achieving efficient video object segmentation.
However, in more complex scenes, such as those depicted in Figure~\ref{fig:limitation}, it becomes challenging to distinguish background distractions from the target due to their high similarity in appearance and location.
% However, accurately accomplishing ID propagation and segmentation in scenes with objects highly similar to the target and occlusions remains challenging for our approach.
%
To be more specific, Figure~\ref{fig:limitation} shows that our method erroneously propagates the IDs when targets 1 and 2 are obscured by similar objects.
This may be due to the severe loss of discriminative features when the target is occluded, leading to issues of ID-switch and difficulty in correctly distinguishing between the target and background distractions.
A potential direction to improve this issue could be incorporating additional information, such as motion information or textual descriptions, into the modeling process.

\section{Conclusion}
In this paper, we present a spatial-temporal multi-level association framework to achieve efficient video object segmentation.
Our proposed method enhances efficiency while ensuring sufficient interaction of feature information by separating the tasks of feature extraction and interaction modeling into three distinct components: object self-attention, reference target enhancement, and target object correlation.
In the spatial-temporal multi-level feature association module, we enhance the feature information related to the multiple targets through asymmetric interactions involving three types of information.
Then, we separate the features of each target through ID propagation from the enhanced feature map for decoding.
Our methodology can be described as the augmentation of features for multiple targets, followed by assigning unique identification to each target.
The favorable performance of all the datasets against the state-of-the-art methods demonstrates the effectiveness of the proposed algorithm.
\clearpage  % TODO REVIEW/FINAL: This \clearpage needs to be removed from both review and camera-ready versions.

% ---- Bibliography ----
%
% BibTeX users should specify bibliography style 'splncs04'.
% References will then be sorted and formatted in the correct style.
%
\bibliographystyle{splncs04}
\bibliography{main}
\end{document}